\documentclass[graybox]{svmult}

\usepackage{type1cm}        
 \usepackage{ulem}                            
\usepackage{makeidx}         
\usepackage{graphicx}        
\usepackage{multicol}        
\usepackage[bottom]{footmisc}

\usepackage{newtxtext}       %
\usepackage[varvw]{newtxmath}       

\usepackage[capitalise]{cleveref}

\definecolor{bostonuniversityred}{rgb}{0.8, 0.0, 0.0}
\definecolor{darkspringgreen}{rgb}{0.09, 0.45, 0.27}
\definecolor{ferrarired}{rgb}{1.0, 0.11, 0.0}
\definecolor{coralred}{rgb}{1.0, 0.25, 0.25}
\definecolor{pastelred}{rgb}{1.0, 0.41, 0.38}
\definecolor{darkblue}{RGB}{0,0,139}
\definecolor{darkred}{RGB}{105,0,0}

\usepackage{tikz}
\usepackage{pgfplots}
\usetikzlibrary{matrix}
\usetikzlibrary{shapes,patterns,snakes,arrows}
\usetikzlibrary{positioning,fit,calc}
\usetikzlibrary{graphs}

\tikzset{%
  every neuron/.style={
    circle,
    draw,
    minimum size=0.6cm
  },
  every input neuron/.style={
    circle,
    draw,
    minimum size=0.6cm,
    fill=green!50
  },
  every output neuron/.style={
    circle,
    draw,
    minimum size=0.6cm,
    fill=orange!30
  },
  every hidden neuron/.style={
    circle,
    draw,
    minimum size=0.6cm,
    fill=blue!40
  },
  neuron missing/.style={
    draw=none, 
    scale=1.5,
    text height=0.3cm,
    execute at begin node=\color{black}$\vdots$
  },
}

\makeindex             

\begin{document}

\title*{Domain-decomposed image classification algorithms using linear discriminant analysis and convolutional neural networks} 
\titlerunning{Domain-decomposed image classification algorithms using LDA and CNNs} 

\author{Axel Klawonn\orcidID{0000-0003-4765-7387} \\ Martin Lanser\orcidID{0000-0002-4232-9395} \\ Janine Weber\orcidID{0000-0002-6692-2230}}
\institute{Axel Klawonn, Martin Lanser, Janine Weber \at Department of Mathematics and Computer Science, University of Cologne, Weyertal 86-90,\\ 50931 K\"oln, Germany, \url{https://www.numerik.uni-koeln.de}\\
Center for Data and Simulation Science, University of Cologne, 50923 K\"oln, Germany, \url{https://www.cds.uni-koeln.de}\\  \email{\{axel.klawonn,martin.lanser,janine.weber\}@uni-koeln.de}}
%
%
\maketitle

\abstract{In many modern computer application problems, the classification of image data plays an important role. Among many different supervised machine learning models, convolutional neural networks (CNNs) and linear discriminant analysis (LDA) as well as sophisticated variants thereof are popular techniques. In this work, two different domain decomposed CNN models are experimentally compared for different image classification problems. Both models are loosely inspired by domain decomposition methods and in addition, combined with a transfer learning strategy. The resulting models show improved classification accuracies compared to the corresponding, composed global CNN model without transfer learning and besides, also help to speed up the training process. 
Moreover, a novel decomposed LDA strategy is proposed which also relies on a localization approach and which is combined with a small neural network model. In comparison with a global LDA applied to the entire input data, the presented decomposed LDA approach shows increased classification accuracies for the considered test problems. 
}

\section{Introduction}
\label{sec:intro}

In many modern computer application problems, the classification of image data, or, more general, of data with a grid-like structure, plays an important role. Common examples are face recognition~\cite{parkhi:2015:deepface}, medical image diagnosis~\cite{singh:2020:3d_med_review}, or general object detection~\cite{lecun1989backpropagation}. 
Among many different machine learning models, in particular convolutional neural networks (CNNs)~\cite{lecun:1989:CNN} have been shown to be very successful in approximating image classification tasks. 
However, with increasing numbers of model parameters and increasing availability of large amounts of training data, the training of CNNs requires high memory availability and results in long training times. 
Here, parallelization approaches for a time- and memory-efficient training process of large CNNs have become increasingly important; see also~\cite{ben2019demystifying} for an overview and a categorization into model and data parallel methods. 

From an abstract point of view, many model parallel training approaches can be interpreted as domain decomposition methods (DDMs)~\cite{toselli}; see~\cite{KLW:DD_ML_survey:2023} for a survey of existing work based on the combination of machine learning and DDMs.
Generally speaking, the main idea is to decompose a large global problem into a number of smaller subproblems which can then be solved in parallel and each of them with less computational effort than the entire global problem. In~\cite{KLW:DNN-CNN:2023,KLW:CNN-DNN_coh:2024,GuCai:2022:dd_transfer,verburg2024ddu}, different model parallel approaches inspired by DDMs have been proposed for the training of different CNN architectures. 
In this paper, we extend our work from~\cite{KLW:DNN-CNN:2023,KLW:CNN-DNN_coh:2024} and compare the model parallel training approach for CNNs which is combined with transfer learning~\cite{KLW:CNN-DNN_coh:2024} with the related approach from~\cite{GuCai:2022:dd_transfer}. This is done for three different datasets and two different CNN architectures within a common implementation framework.

Besides the parallelization of the training process of large CNNs, or neural networks, in general, dimensionality reduction also plays an important role within fast machine learning applications. Linear discriminant analysis (LDA)~\cite{fisher1936use} provides such a tool for extracting the most discriminant features of a dataset and, additionally, for the supervised classification of the reduced low-dimensional data. 
In~\cite{le2020LDA_brain}, LDA has been applied to polyethylene terephthalate (PET) brain images for the classification of patients with Alzheimer's disease and in~\cite{tougaccar2020LDA_CNN}, a combination of LDA and CNNs is used for the classification of invasive breast cancer images, to just mention a few examples where LDA is used for image classification. 
Often, the classic LDA performs well for classification tasks with a small number of classes or features but deteriorates when the number of classes is large; see also~\cite{LiCai:2024:ddLDA}. Hence, in~\cite{LiCai:2024:ddLDA}, an approach has been proposed which combines LDA with DDM to efficiently solve classification problems with a large number of classes. This approach is based on a decomposition of the data into subsets with a smaller number of classes.
In this work, we present a novel approach which combines LDA with DDMs in a different manner and which is directly analogous to our CNN-DNN approach from~\cite{KLW:DNN-CNN:2023}. Here, ideas from DDMs are used to decompose the input images into smaller subimages and for each of the subimages, a local LDA is applied separately. The obtained local classification probability values are then combined with the help of a dense neural network (DNN) which is trained with respect to the global image classification task. 

The remainder of the paper is organized as follows. In~\cref{sec:cnn_lda}, we briefly describe the mathematical ideas behind CNNs and LDA used for image classification problems. In~\cref{sec:models}, we introduced three domain decomposed image classification models of which two are based on CNN models and the third uses LDA. We present experiments for different image classification problems for all three approaches in~\cref{sec:results} and compare the results within a common framework. Finally, in~\cref{sec:concl}, we formulate concluding remarks and potential topics for further future research.

\section{Image classification with convolutional neural networks and linear discriminant analysis}
\label{sec:cnn_lda}

CNNs and LDA are both supervised machine learning models which can both be applied to image classification problems. While CNNs make use of local convolutions to stepwise extract local features of an input image, LDA extracts the most discriminant features of a dataset and hence, is often also used for the purpose of dimensionality reduction. In this section, we provide a brief mathematical description of the two supervised machine learning models.

\subsection{Convolutional neural networks}
\label{sec:cnn}

CNNs are a particular type of neural networks which are specialized for processing data with a grid-like structure such as, for example, image data in form of pixel or voxel grids, or time series data. 
The following general description of CNNs is roughly based on~\cite[Chapt. 9]{Goodfellow:2016:DL},~\cite[Chapt. 5]{chollet2017deep}, and~\cite[Sect. 2]{KLW:DNN-CNN:2023} and can be skipped by the experienced reader already familiar with CNNs.
 
The general structure of a CNN can basically be described by two types of layers, that is, convolutional layers and pooling layers. Both are a specific kind of linear operations.
Convolutional layers are mathematically motivated by discrete convolutions of two discrete functions. 
Such a discrete convolution can be written as a multiplication of a tensor with a \textit{kernel} matrix and basically corresponds to the computation of a weighted average function. The concrete weight entries in form of the kernel are learned during the network training. For most CNNs used in practical applications, the kernel has usually a much smaller size than the input data and hence, CNNs typically have sparse interactions and are very effective at extracting local, meaningful features from the input data; see~\cite[Chapt. 9]{Goodfellow:2016:DL} and~\cite[Chapt. 5]{chollet2017deep}.
The complete output of a convolutional layer is then obtained by sliding the kernel over patches of fixed size of the input data while successively computing the weighted sum of the kernel entries and the corresponding values along the patches. The obtained output of a convolutional layer is typically referred to as \textit{feature map}. In general, a CNN usually uses multiple kernels or \textit{filters} on subsequent feature maps one after another such that the input data are compressed layer by layer.

Besides by the size of its kernel, a convolutional layer is also characterized by the choice of \textit{padding} and \textit{striding} it uses. 
Padding basically adds an appropriate number of rows and columns filled with zero entries on each side of the input feature map such that it is possible to fit the kernel windows also around the border patches of the original input feature map and to preserve the size of the input data within the following feature maps.
Otherwise, that is, without the use of padding, the output feature maps will shrink with each convolutional layer given that you cannot fully center the kernel around the border patches of the feature map.
Additionally, striding defines the shift of the local kernels within the convolutions and hence, also determines the reduction of the data dimension within a convolutional layer. For more details on different types of padding and striding, see also~\cite[Chapt. 9]{Goodfellow:2016:DL} and~\cite[Chapt. 5]{chollet2017deep}.
Let us note that, usually, a stack of convolutional layers is followed by a nonlinear activation function in order to learn a more general nonlinear functional relation between the input and the output data of the CNN.

Typically, a stack of convolutional layers is followed by a \textit{pooling layer}. A pooling layer replaces the output within a patch of fixed size of a given feature map by a summary statistic of the underlying values.  Common examples for such pooling operations are a maximum pooling or an average pooling. The latter replaces a rectangular neighborhood with its average value whereas the former replaces it with its maximum value. Both variants help to further downsample the respective feature map and make it more invariant to small translations of the input; see also~\cite[Chapt. 9]{Goodfellow:2016:DL}.

\subsection{Linear discriminant analysis}
\label{sec:lda}

Besides neural networks, LDA is a further popular technique for supervised classification problems. 
Generally speaking, the main objective of LDA is to identify the most discriminant features of a dataset and to project the original data onto these features such that the data can be well separated into nonoverlapping classes within the reduced low-dimensional feature space. For the readers not already familiar with LDA, we provide a short description of the mathematical background in this section.
The following descriptions are roughly based on~\cite[Chapt. 5]{brunton2022data},~\cite[Chapt. 4]{bishop2006pattern}, and~\cite[Sect. 3]{LiCai:2024:ddLDA}.

Let us assume that we have a high-dimensional data matrix $X \in \mathbb{R}^{m\times n}$ with datapoints belonging to $c$ classes where $x_{ij} \in \mathbb{R}^m$ denotes the $i$-th sample in the $j$-th class. Each of the $c$ classes contains $n_j, j=1,\ldots,c,$ data points and $n=\sum_{j=1}^c n_j$ equals the total number of samples in the dataset. 
We denote by 
\begin{equation}
\mu = \frac{1}{n} \sum_{j=1}^c \sum_{i=1}^{n_j} x_{ij}
\end{equation}
the mean of the entire dataset $X$ and by
\begin{equation}
\mu_j = \frac{1}{n_j} \sum_{i=1}^{n_j} x_{ij}
\end{equation}
the mean of the $j$-th class.
Then, we define the \textit{within-class scatter matrix} $S_W$ by
\begin{equation}
S_W = \sum_{j=1}^c \sum_{i=1}^{n_j} (x_{ij} - \mu_j) (x_{ij} - \mu_j)^T
\end{equation}
as well as the \textit{between-class scatter matrix} $S_B$ by
\begin{equation}
S_B = \sum_{j=1}^c n_j (\mu_j - \mu)(\mu_j - \mu)^T.
\end{equation}

The LDA now computes a low-dimensional projection matrix $V\in \mathbb{R}^{m\times d}$ by maximizing the between-class variance while simultaneously minimizing the within-class variance in the projected space. Here, $d$ denotes the desired dimension of the reduced feature space and, usually, we aim for $d\ll n$.
Mathematically, the LDA computes the orthogonal matrix $V$ that maximizes the ratio
\begin{equation}
\max_V J(V) = \frac{\text{Tr}(V^T S_B V)}{\text{Tr}(V^T S_W V)}
\end{equation}
where $\text{Tr}$ denotes the trace of a matrix. 
As shown in~\cite{fukunaga2013intro}, the optimal projection $V^*$ can be computed as the set of eigenvectors $v_i \in \mathbb{R}^m$ of the generalized eigenvalue problem
\begin{equation}
S_B v_i = \lambda_i S_W v_i
\end{equation}
corresponding to the largest $d$ eigenvalues and hence, we obtain $V^* = [v_1, v_2, \ldots, v_d] \in \mathbb{R}^{m \times d}$.
For more details of the computation of the optimal low-dimensional projection matrix $V^*$, we refer to~\cite[Chapt. 4.1]{bishop2006pattern}.

\section{Domain decomposed image classification models}
\label{sec:models}

In this section, we present different approaches that aim at a model parallel training of different image classification machine learning models.
In~\cref{sec:CNN-DNN} and~\cref{sec:transfer_cai}, we present two approaches for a model parallel training of CNN architectures. Both approaches are inspired by domain decomposition methods~\cite{toselli} and make use of a transfer learning strategy. However, the global, composed CNN model differs for both approaches. Finally, in~\cref{sec:decomp_lda}, we present a novel approach which applies the idea of our CNN-DNN model from~\cite{KLW:DNN-CNN:2023} to LDA, that is, LDA is used for the supervised image classification instead of CNNs.

\subsection{Coherent CNN-DNN model architecture}
\label{sec:CNN-DNN}

\begin{figure}
\centering
\includegraphics[width=0.9\textwidth]{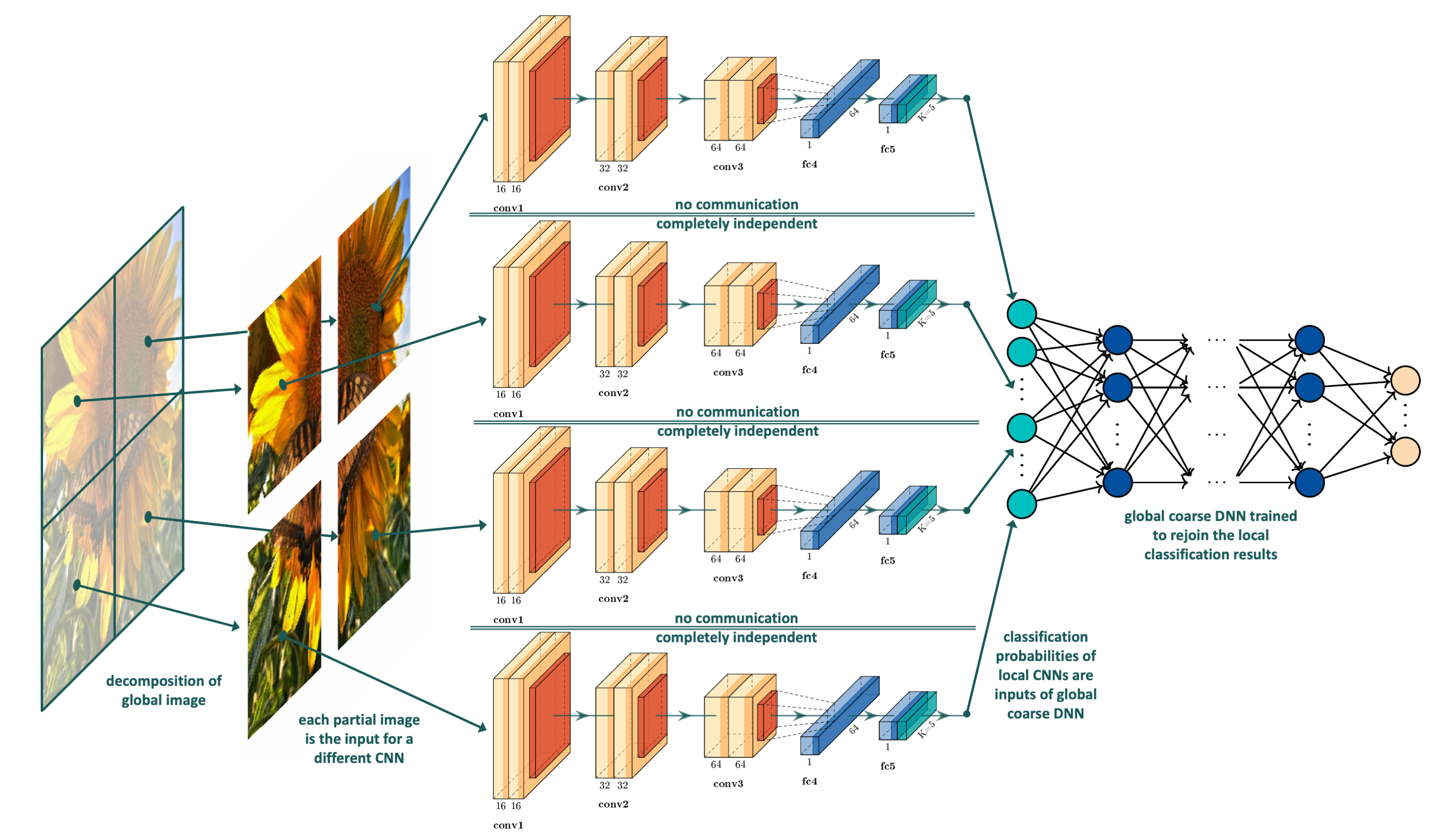}
\caption{Visualization of the CNN-DNN network architecture. \textbf{Left:} The original image is decomposed into $N=4$ nonoverlapping subimages. \textbf{Middle:} The $N=4$ subimages are used as input data for $N$ independent, \textit{local} CNNs. \textbf{Right:} The probability values of the local CNNs are used as input data for a DNN. The DNN is trained to make a final classification for the decomposed image by weighting the local probability distributions.
Figure taken from~\cite[Fig. 4]{KLW:DNN-CNN:2023}.}
\label{fig:dd_cnn}
\end{figure}

In~\cite{KLW:DNN-CNN:2023}, a novel hybrid CNN-DNN architecture is proposed which is loosely inspired by domain decomposition and which naturally supports a model parallel training strategy. This network architecture is defined by a spatial decomposition of the input images into smaller subimages. As a starting point, we assume that we have a global CNN model that takes as input data a two-dimensional pixel image with $H\times W$ pixels and outputs a probability distribution with respect to $K \in \mathbb{N}$ classes. For the definition of the CNN-DNN model from~\cite{KLW:DNN-CNN:2023}, we decompose the input images into a finite number of $N \in \mathbb{N}$ smaller subimages. Let us note that for colored images with $3$ channels of $H\times W$ pixels, we exclusively decompose the images in the first two dimensions, the height and the width, but not in the third dimension. Hence, each image is decomposed into $N$ subimages with height $H_i$ and width $W_i, \ i=1, \ldots, N$. 
Then, for each of these subimages, proportionally smaller CNNs are trained that operate exclusively on certain subimages of all input images. In particular, these smaller CNNs can be trained in parallel and due to the analogy to DDMs, we denote these CNNs as local CNNs. 
 All local CNNs are defined such that they always have the same general structure as the original global CNN but differ in the number of channels of the feature maps, the number of neurons within the fully connected layers, as well as in the number of input nodes. All of these layers are proportionally smaller than for the respective global CNN.  As output data for each of the local CNNs, we obtain a set of $N$ local probability distributions with respect to the $K$ classes, where each of the local probability distributions corresponds to a local decision exclusively based on information extracted from the local subimages. Let us note that the described decomposition of the input data can be applied completely analogously to three-dimensional input data, that is, voxel data. In such cases, the respective CNN model consists of three-dimensional instead of two-dimensional convolutional and pooling operations. 
As originally proposed in~\cite{KLW:DNN-CNN:2023}, after the training of the local CNNs, a feedforward dense neural network (DNN) is trained which evaluates the local probability distributions into a final, global decision with respect to the given image classification problem. This results in a hybrid CNN-DNN model architecture. Hence, the DNN takes as input data a vector containing the $K \ast N$ local probability values of all $N$ local CNNs and is trained to map this input vector to the correct classification labels of the original input images. 
An exemplary visualization of this CNN-DNN model for a global CNN of VGG3~\cite{simonyan:2014:VGGnet} type is shown in~\cref{fig:dd_cnn}.

In~\cite{KLW:CNN-DNN_coh:2024}, a modified version of the described model parallel training approach has been proposed and investigated with respect to the classification accuracies for different image recognition problems. In particular, for the experiments presented in~\cite{KLW:CNN-DNN_coh:2024}, the CNN-DNN is sequentially trained as one coherent  model. That means that the CNN-DNN architecture as shown in~\cref{fig:dd_cnn} is implemented as one connected model using the functional API of Tensorflow and trained within one sequential training loop. For the remainder of this paper, we refer to this model as \textit{coherent CNN-DNN} (CNN-DNN-coherent). 
Additionally, as also investigated in~\cite{KLW:CNN-DNN_coh:2024}, the training of the coherent CNN-DNN can be combined with a transfer learning strategy. In this case, we first train the local CNNs on separate subimages for a fixed number of epochs and subsequently use the obtained network parameters as initializations for the respective weights and bias values of the coherent CNN-DNN model. The coherent CNN-DNN with this pretrained initialization is then trained further with respect to the global image classification problem. For the experiments in~\cite{KLW:CNN-DNN_coh:2024}, using the coherent CNN-DNN model in combination with a transfer learning strategy leads to improved classification accuracies for both, a VGG~\cite{simonyan:2014:VGGnet} and a residual network (ResNet)~\cite{he2016deep} model architecture, compared to the training strategies without transfer learning. Hence, in~\cref{sec:results}, we focus on the experimental comparison of the coherent CNN-DNN model with transfer learning, even if it comes at the cost of reduced parallelization potential.  In the following, we refer to this approach as \textit{CNN-DNN with transfer learning} (CNN-DNN-transfer).

\subsection{Decomposed CNN with transfer learning}
\label{sec:transfer_cai}

A further method to parallelize the training of large CNNs in combination with transfer learning has been proposed in~\cite{GuCai:2022:dd_transfer}.  
Here, the authors propose to decompose a global CNN into several smaller sub-networks by decomposing the width, that is, the channel dimension while keeping the depth of the CNN constant.
Analogously to the approach presented in~\cref{sec:CNN-DNN}, this decomposition also corresponds to a decomposition of the input images into $N \in \mathbb{N}$ rectangular subimages.
All local sub-networks are trained in parallel and independently from each other while operating exclusively on the respective subimages of the input data. 
This model parallel training is then combined with a transfer learning strategy, where the weights of the trained sub-networks are used to initialize the weights of the global network. In case that also weights connecting different sub-networks exist, these weights are set as zero tensors. The global CNN is then trained for a given number of epochs to further adjust the network parameters with respect to the global image classification problem. 
Within the context of domain decomposition, the described procedure can also be interpreted as a nonlinear preconditioning strategy; see also~\cite{GuCai:2022:dd_transfer}.
Let us note that both approaches from~\cref{sec:CNN-DNN} and~\cref{sec:transfer_cai} are based on a decomposition of the input images into smaller subimages and in both cases, smaller CNN models are trained in parallel. However, the composed global CNN model differs for both approaches. In particular, for the approach introduced in~\cite{GuCai:2022:dd_transfer}, the global CNN model needs always to be trained as a whole after the pretraining of the smaller local CNNs, even though for a smaller number of epochs compared to the training without a transfer learning strategy.   
For the remainder of the paper, we denote this approach as \textit{domain decomposed CNN with transfer learning} (DD-CNN-transfer).

Moreover, the authors of~\cite{GuCai:2022:dd_transfer} provide a theoretical analysis to estimate the effectiveness of their model parallel transfer learning strategy. More precisely, a theoretical proof is presented showing that the pretrained local, smaller CNN models provide an upper and a lower bound for the cost function and the classification accuracy of the corresponding global CNN model; cf.~\cite[Theorem 2.2]{GuCai:2022:dd_transfer} for more details.

\subsection{Decomposed LDA}
\label{sec:decomp_lda}

In this section, we present a novel approach which applies the idea of our CNN-DNN machine learning model from~\cite{KLW:DNN-CNN:2023} to LDA. This means that we aim for a model parallel method for image recognition problems which uses LDA for the supervised image classification instead of CNNs. 
Analogously to the approach presented in~\cref{sec:CNN-DNN}, the novel approach is based on a decomposition of the input data in form of images into $N$ nonoverlapping subimages with height $H_i$ and width $W_i, \ i=1, \ldots, N$. For each of these subimages, we then compute local low-dimensional feature spaces based on local projections onto the most discriminant features and define linear decision boundaries to separate the reduced low-dimensional data into the underlying classes. This results in a local probability distribution among the $K$ classes of the underlying classification problem for each of the $N$ subimages, similar to~\cref{sec:CNN-DNN} where the local probability distributions are obtained from local CNNs.
In order to combine the local classifications into a final global decision, we again subsequently train a DNN that uses as input data all local probability distributions collected in one vector with $K \ast N$ entries and outputs a final, global probability distribution. Due to the analogy to the hybrid approach presented in~\cref{sec:CNN-DNN}, we denote the novel approach by \textit{LDA-DNN} for the remainder of the paper.  

Let us note that in~\cite{LiCai:2024:ddLDA}, also a domain decomposed classification algorithm based on LDA was proposed. However, the approach presented in~\cite{LiCai:2024:ddLDA} relies on a different decomposition strategy such that the training data are separated in disjunct subsets of classes. In particular, for training data with a large number of classes, the decomposition from~\cite{LiCai:2024:ddLDA} results in local subsets where each subset only includes a smaller number of classes. 
Additionally, the authors of~\cite{LiCai:2024:ddLDA} use a predefined strategy to combine the local projection spaces obtained from the decomposed LDAs to obtain a final global classification for unseen test data. More precisely, they compute a scaled distance vector with respect to the center of all classes and minimize the scaled distance between the projected test data and the respective center of all classes. 
 This is also different from our new LDA-DNN approach where the global decision strategy is learned automatically by a neural network.  
A more detailed theoretical as well as an experimental comparison of both approaches is a potential topic of future research.

\section{Experiments}
\label{sec:results}

In the following, we present some experiments with respect to the described approaches from~\cref{sec:models} and compare the classification accuracies for different image classification datasets. All experiments have been carried out on a workstation with 8 NVIDIA Tesla V100 32GB GPUs using the TensorFlow library.

\subsection{Network architectures and datasets}

\begin{figure}[t]
\centering
\includegraphics[width=0.35\textwidth]{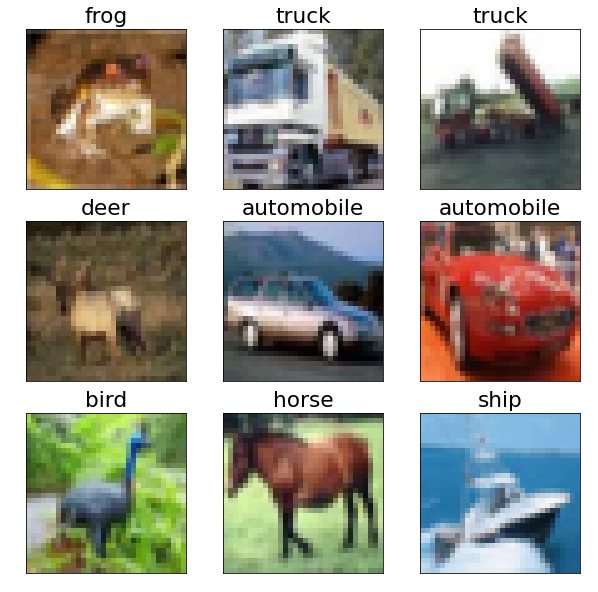}
\hspace{1cm}
\includegraphics[width=0.35\textwidth]{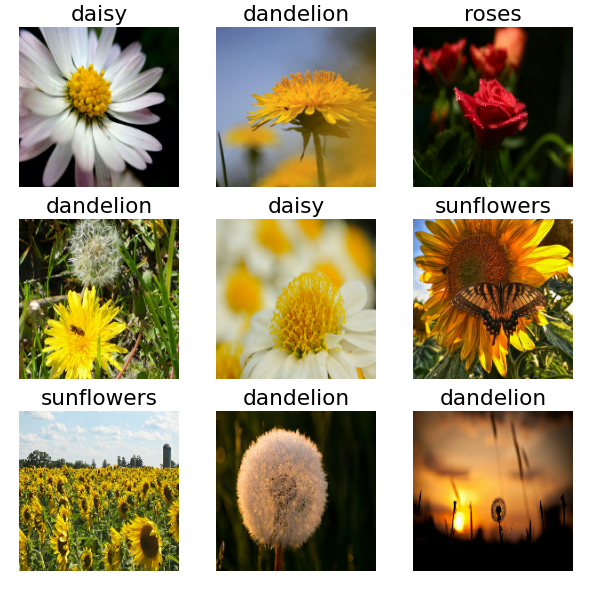}
\caption{\textbf{Left:} Exemplary images of the CIFAR-10 dataset~\cite{Cifar10_TR}. \textbf{Right:} Exemplary images of the TF-Flowers dataset~\cite{tfflowers}.}
\label{fig:cifar10_ex}
\end{figure}

\begin{figure}[t]
\centering
\includegraphics[width=0.65\textwidth]{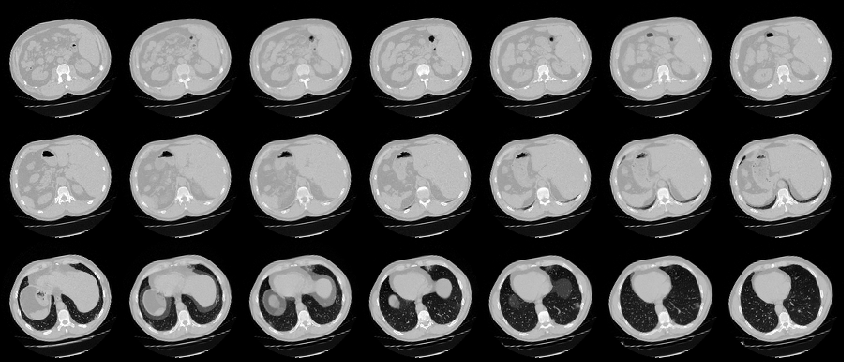}
\caption{Exemplary slices for one chest CT scan taken from the MosMedData dataset~\cite{Chest_CT}.}
\label{fig:CT_lung_ex}
\end{figure}
 
To evaluate and compare the performance of the introduced domain decomposed image classification models from~\cref{sec:models}, we consider three different datasets and, for the CNN based approaches, two different network architectures.

First, we test all three approaches from~\cref{sec:models} for two different two-dimensional image datasets, that is, the CIFAR-10 data~\cite{Cifar10_TR} and the TF-Flowers data~\cite{tfflowers}.
The CIFAR-10 dataset is often used for benchmarking purposes of new machine learning models and consists of $50\,000$ training and $10\,000$ validation images of $32\times 32$ pixels each. All images are categorized into $K=10$ different classes, for example, different animals or different vehicles; see also~\cref{fig:cifar10_ex} (left).
The TF-Flowers dataset~\cite{tfflowers} consists of $3\,670$ images of $180\times 180$ pixels which are categorized into $K=5$ different classes of flowers; cf. also~\cref{fig:cifar10_ex} (right). We split these images into $80\%$ training and $20\%$ validation data.
Additionally, we further test all three approaches for a three-dimensional dataset of chest CT scans~\cite{Chest_CT}. This dataset consists of CT scans with and without signs of COVID-19 related pneumonia, that is, we have $K=2$ classes. Each of these CT scans consists of $128\times 128 \times 64$ voxels and hence, for this dataset, we train CNN models using three-dimensional filters and pooling layers.
An exemplary visualization of CT slices for one exemplary datapoint can be seen in~\cref{fig:CT_lung_ex}.

For the two domain decomposed image classification models based on CNNs from~\cref{sec:CNN-DNN} and~\cref{sec:transfer_cai}, we consider two different CNN architectures. As a first test, we consider a CNN with nine blocks of stacks of convolutional layers and a fixed kernel size of $3\times 3$ pixels, in case of two-dimensional image data, or $3\times 3 \times 3$ voxels, for three-dimensional image data, respectively. We refer to this network architecture as VGG9 for the remainder of this paper and refer to~\cite{simonyan:2014:VGGnet} for more details of this network model. 
As a second test, we implement a ResNet~\cite{he2016deep} with 20 blocks of convolutional layers where we add skip connections between each block and its third subsequent block; see also~\cite{he2016deep} for more technical details. We refer to this network architecture as ResNet20.  
Both networks are trained using the Adam (Adaptive moments) optimizer~\cite{Kingma:2014:MSO} with its default parameters and the cross-entropy loss function. 
For the DNN, which is included in the CNN-DNN approach, we use a dense feedforward neural network with four hidden layers and $\{128, 64, 32, 10 \}$ neurons within the hidden layers. This architecture has been optimized by a gridsearch; see also~\cite{KLW:DNN-CNN:2023}.

For the LDA-DNN approach, we select $d=2$ for the number of considered discriminant vectors for both, the TF-Flowers and the CIFAR-10 data. For the chest CT data, we use $d=1$ since in this case, the LDA algorithm needs to differentiate between $K=2$ classes exclusively. Let us note that, in general, a fixed choice of the dimension parameter $d$ for all local LDAs may not be the optimal choice. More sophisticated strategies might exist, as, for example, choosing higher values of $d$ for local LDAs which are computed for a set of subimages which cover a higher number of classes and using a lower value of $d$ for a set of subimages that cover a smaller number of classes, respectively. However, for a first proof of concept whether the proposed LDA-DNN approach leads to promising classification accuracies, a constant choice of $d$ for all considered subspaces seems to be sufficient.

\subsection{Results and discussion}

In~\cref{tab:ddCNN_LDA_acc}, we compare the classification accuracies for three different domain decomposed image classification models as introduced in~\cref{sec:models}.
The CNN-DNN-transfer and the DD-CNN-transfer model follow a similar strategy in the sense that they first train local, proportionally smaller CNNs operating on local subimages and use the obtained network parameters as the initialization for a global hybrid CNN-DNN or global CNN model, respectively, which is then trained further. In both cases, we first train the local CNNs for $150$ epochs and subsequently train the global network with the specific initialization for additional $50$ epochs. 
As we can observe from~\cref{tab:ddCNN_LDA_acc}, we obtain similar results in terms of classification accuracy for the validation data for the CNN-DNN-transfer and the DD-CNN-transfer model for the CIFAR-10 and the TF-Flowers dataset for all tested decompositions of the input images. For the chest CT data, we obtain only slightly lower accuracies values for the DD-CNN-transfer model compared to the CNN-DNN-transfer approach. 
This observation is fairly interesting since the coherent global model is different in both cases. For both models, using a transfer learning strategy seems to be a suitable method to achieve high classification accuracies which are comparable or even higher than for the respective global CNN model which was trained without any model parallelism and transfer learning; cf. also the corresponding accuracy values in the first column of~\cref{tab:ddCNN_LDA_acc}. 

In the last column of~\cref{tab:ddCNN_LDA_acc}, we report the classification accuracies for the LDA-DNN approach for the same three datasets as before. Let us note that for the LDA-DNN approach, we have no analogous differentiation which local classification model is used as for the different local CNN network architectures (VGG and ResNet), since, for each dataset, we first compute the $d$ most discriminant features and use these features to project both the training as well as the validation data onto the reduced low-dimensional subspace defined by the identified discriminants. 
As we can observe from~\cref{tab:ddCNN_LDA_acc}, the LDA-DNN approach leads to satisfying classification accuracies for all three test datasets which are comparable to the results for the global CNN model (first column of~\cref{tab:ddCNN_LDA_acc}). Additionally, the LDA-DNN shows an improved performance compared to a global LDA which is computed for the entire input images. 
However, the percentage figures for the LDA-DNN are lower than for the two considered decomposed CNN approaches. Additionally, the LDA-DNN shows a stronger tendency of overfitting since for all three datasets, the accuracy values with respect to the training data are about $20\%$ higher than for the validation data. This indicates that the chosen number $d$ of most relevant discriminants might not be optimal for the considered local data or, more general, that the computed rather drastical dimensionality reduction of the images is only partially suitable to provide good generalization properties. More detailed investigation of this effect, also from a theoretical point of view, is a potential topic for future research. 

\begin{figure}
\centering
\scalebox{0.88}{
\begin{tikzpicture}
\begin{axis} [xbar stacked,width=11cm,height=5cm, axis y line*=none,
    axis x line*=bottom, ytick={0,1,2,3,4}, yticklabels={DD-CNN-transfer $4\times 4$, DD-CNN-transfer $2\times 2$, CNN-DNN-transfer $4\times 4$, CNN-DNN-transfer $2\times 2$, Global}, xlabel = Training time in s, xmajorgrids=true, xmin=0, xmax=100, xtick={10,20,30,40,50,60,70,80,90,100}, enlarge x limits=0.0, legend style={legend columns=5,at={(0.45,1.25)},anchor=north,draw=none, font=\small}]
\addplot+[xbar,gray] coordinates { 
(0,0)
(0,1)
(0,2)
(0,3)
(89.89,4)
};
\addplot+[xbar,darkblue] coordinates { 
(13.55, 0)
(21.35, 1)
(14.66, 2)
(22.10, 3)
(0,4)
};
\addplot+[xbar,pastelred] coordinates { 
(25.55, 0)
(27.35, 1)
(33.27, 2)
(34.79, 3)
(0,4)
};
\legend{Global CNN, Local CNN, Transfer, Local CNN, Transfer}
\end{axis}
\end{tikzpicture}
}
\caption{Comparison of the training times for a global VGG9 model, and the two decomposed CNN models, that is, the CNN-DNN-transfer approach~\cite{KLW:CNN-DNN_coh:2024} and the DD-CNN-transfer approach~\cite{GuCai:2022:dd_transfer} for the TF-Flowers dataset~\cite{tfflowers}. We show in gray the training time required for the global VGG9 model using the entire TF-Flowers images as input data. In blue, we show the training time for the local CNNs for CNN-DNN-transfer and DD-CNN-transfer, respectively, and in red the times for the subsequent training of the respective global net using the transfer learning strategy. }
\label{fig:runtime_flowers_VGG9}
\end{figure}
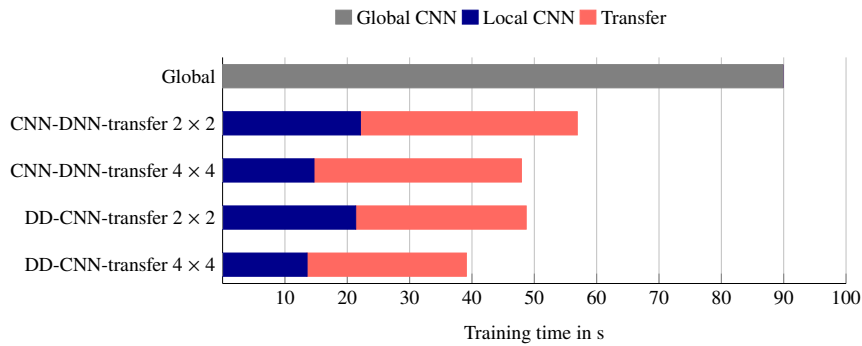

Given that the CNN-DNN-transfer and the DD-CNN-transfer approach are alike in the sense that both are based on decomposed CNNs and show similar classification accuracies in~\cref{tab:ddCNN_LDA_acc}, we additionally compare the training times for both models for the TF-Flowers data. In~\cref{fig:runtime_flowers_VGG9}, we report the training time for a global VGG9 model which is trained on the entire images as input data. Besides, we show in blue the training times for the local CNNs of CNN-DNN-transfer and DD-CNN-transfer as well as in red the times for the subsequent training of the respective global net using the transfer learning strategy. Note that for both methods, the local CNNs are trained for $150$ epochs and the global network is trained for additional $50$ epochs using the obtained network parameters from the local CNNs as initializations. 
As we can observe from~\cref{fig:runtime_flowers_VGG9}, both the CNN-DNN-transfer and the DD-CNN-transfer method result in a reduction of the training time of a factor of approximately $1.57$ to $2.3$ compared to the training of the global model without transfer learning. However, a further reduction in training time is limited by the training of the entire global network model for $50$ epochs. A stronger reduction in training time could, for example, be obtained by further reducing the number of epochs for which the global model is trained such that the training is stopped adaptively once no relevant improvement of the classification accuracy is observed. While our aim in this work was to provide a fair comparison in the sense that all models are trained for the same total number of $200$ epochs and to predominantly compare the classification accuracies, the above mentioned strategy is an interesting topic for future research.  
Finally, let us note that we see a higher potential to reduce the training time of large CNN models for the classification of high-resolution, three-dimensional data; see also~\cite{KLW:DNN-CNN:2023} for first results. 

\begin{table}[t]
\centering
\caption{\label{tab:ddCNN_LDA_acc} Classification accuracies for the validation and training data (in brackets) for a global CNN benchmark model (VGG9 or ResNet20), a coherent CNN-DNN model trained with a transfer learning approach (CNN-DNN-transfer) as introduced in~\cite{KLW:CNN-DNN_coh:2024}, the decomposed CNN model with transfer learning from~\cite{GuCai:2022:dd_transfer} (DD-CNN-transfer), and a decomposed LDA approach (LDA-DNN).}
\scalebox{0.86}{
\begin{tabular}{l||c|c|c||c|c}
    {\bf Decomp.} & {\bf global CNN} & {\bf CNN-DNN-transfer} & {\bf DD-CNN-transfer} & {\bf global LDA} & {\bf LDA-DNN} \\\hline \hline
    \multicolumn{4}{c||}{\bf CIFAR-10, VGG9} & \multicolumn{2}{c}{\bf CIFAR-10}  \\\hline
type A &  0.7585  & {\bf 0.8462} & 0.8403 & 0.5620 & 0.7557 \\
$2\times 2 $, $\delta=0$ &  (0.8487)  & (0.8889) & (0.8767) & (0.9074) & (0.9434) \\\hline
\multicolumn{4}{c}{\bf CIFAR-10, ResNet20}  \\\hline
type A &  0.8622  & {\bf 0.9117} & 0.9047 & - & - \\
$2\times 2 $, $\delta=0$ &  (0.9343) & (0.9664) & (0.9288) & - & - \\\hline \hline
    \multicolumn{4}{c||}{\bf TF-Flowers, VGG9} & \multicolumn{2}{c}{\bf TF-Flowers}  \\\hline
type A &  0.7887 & {\bf 0.8378} & 0.8289 & 0.5897 & 0.7854 \\ 
$2\times 2,$ $\delta=0$  & (0.9321) & (0.8999) & (0.8773) & (0.8893) & (0.9686) \\\hline
type A  & 0.7887 &  {\bf 0.8608} & 0.8522 & 0.5897 & 0.7643 \\ 
$4\times 4,$ $\delta=0$  & (0.9321)  & (0.8806) & (0.8784) & (0.8893) & (0.9508)\\\hline 
 \multicolumn{4}{c||}{\bf TF-Flowers, ResNet20} & \multicolumn{2}{c}{\bf TF-Flowers}  \\\hline
 type A  & 0.8227 & {\bf 0.8997} & 0.8556 & - & - \\ 
$2\times 2,$ $\delta=0$  & (0.9178)  & (0.9702) & (0.9054) & - & - \\\hline 
type A  & 0.8227 & {\bf 0.8654} & 0.8474 & - & - \\ 
$4\times 4,$ $\delta=0$  & (0.9178) & (0.9244) & (0.8997) & - & - \\\hline \hline
    \multicolumn{4}{c||}{\bf Chest CT scans, VGG9} & \multicolumn{2}{c}{\bf Chest CT scans}  \\\hline
    type A &  0.7667 & {\bf 0.9304} & 0.9066 & 0.6004 & 0.7303 \\
$2\times2\times1$, $\delta=0$  & (0.8214)  & (0.9577) & (0.9449) & (0.8457) &(0.96776)  \\\hline
type A &  0.7667 & {\bf 0.9025} & 0.8888 & 0.6004 & 0.7048\\
$4\times4\times2$, $\delta=0$  & (0.8214) & (0.9488) & (0.9476) & (0.8457) & (0.8998) \\\hline
\end{tabular}
}
\end{table}

\section{Conclusion and future work}
\label{sec:concl}

In this work, we have compared two different decomposed CNN models for the classification of different image datasets. Both decomposed CNN models are loosely inspired by DDMs and are further combined with a transfer learning strategy. Even though the composed global model differs, both, the CNN-DNN-transfer and the DD-CNN-transfer method lead to improved classification accuracies compared to a global CNN model without transfer learning. In our experiments, the CNN-DNN-transfer model results in slightly higher accuracy values. 
Moreover, both methods can help to reduce the required training time of large CNN models. However, the number of epochs for which the global, coherent model needs to be trained after its initialization can become a limiting factor for further acceleration of the training. In future research, a more detailed investigation of an adaptive choice of the number of training epochs for the global model, for example, by using early stopping~\cite{prechelt:1998:early_stop} will be of interest.  

Additionally, we have proposed and investigated a new method that aims at localizing LDA applied to image classification and builds on the work of~\cite{KLW:DNN-CNN:2023}. For the experiments considered in this work, the resulting LDA-DNN approach shows an improved performance compared to a global LDA applied to the entire input images. Simultaneously, the LDA-DNN approach  shows a worse performance in terms of classification accuracies compared to the two CNN-based approaches considered. However, since LDA is a deterministic approach and requires the tuning of less hyperparameters than the optimal design of a neural network, the combination of a decomposed LDA with a small DNN can still be an interesting image classification model which we plan to analyze in more details, also from a theoretical point of view, in future research.

\bibliographystyle{abbrv}
\bibliography{dnn_cnn_smlet24}{}

\end{document}